\documentclass[letterpaper, 10 pt, conference]{ieeeconf}  
\usepackage[cmex10]{amsmath}
\usepackage{amsfonts,amssymb}
 \usepackage{cite}

\usepackage{algorithmic}
\usepackage{algorithm}
\usepackage{graphicx}
\usepackage{subfigure}

\usepackage{booktabs} 
\usepackage{bm}
\usepackage{multirow}

\IEEEoverridecommandlockouts                              

\title{\LARGE \bf
Clustering-Based Subset Selection in Evolutionary Multiobjective Optimization*
}

\author{Weiyu Chen$^{1}$, Hisao Ishibuchi$^{1,2}$, and Ke Shang$^{1}$
\thanks{*This work was supported by National Natural Science Foundation of China (Grant No. 61876075), Guangdong Provincial Key Laboratory (Grant No. 2020B121201001), the Program for Guangdong Introducing Innovative and Enterpreneurial Teams (Grant No. 2017ZT07X386), The Stable Support Plan Program of Shenzhen Natural Science Fund (Grant No. 20200925174447003), Shenzhen Science and Technology Program (Grant No. KQTD2016112514355531).}%
\thanks{$^{1}$ Weiyu Chen, Hisao Ishibuchi, and Ke Shang are with Guangdong Provincial Key Laboratory of Brain-inspired Intelligent Computation, Department of Computer Science and Engineering, Southern University of Science and Technology, Shenzhen 518055, China.
        {\tt\small 11711904@mail.sustech.edu.cn, hisao@sustech.edu.cn, kshang@foxmail.com}}%
\thanks{$^{2}$ Corresponding Author}%
}
\begin{document}

\maketitle
\thispagestyle{empty}
\pagestyle{empty}

\begin{abstract}
Subset selection is an important component in evolutionary multiobjective optimization (EMO) algorithms. Clustering, as a classic method to group similar data points together, has been used for subset selection in some fields. However, clustering-based methods have not been evaluated in the context of subset selection from solution sets obtained by EMO algorithms. In this paper, we first review some classic clustering algorithms. We also point out that another popular subset selection method, i.e., inverted generational distance (IGD)-based subset selection, can be viewed as clustering. Then, we perform a comprehensive experimental study to evaluate the performance of various clustering algorithms in different scenarios. Experimental results are analyzed in detail, and some suggestions about the use of clustering algorithms for subset selection are derived. Additionally, we demonstrate that decision maker's preference can be introduced to clustering-based subset selection.   

\end{abstract}

\section{Introduction}
Evolutionary multiobjective optimization (EMO) algorithms have demonstrated great success in solving multi-objective optimization problems. In the EMO field, subset selection is one of the most important topics, which is involved in many phases of EMO algorithms. 

During environmental selection, a subset of parents and offspring needs to be selected for the next generation. To this end, various subset selection methods have been used, such as dominance and density-based selection (e.g., NSGA-II \cite{NSGA-II}), reference vector-based selection (e.g., MOEA/D \cite{MOEAD}), and indicator-based selection (e.g., SMS-EMOA \cite{SMSEMOA}). 

Additionally, if we run an EMO algorithm with an unbounded external archive (UEA) \cite{HisaoECAI, DSS, BenchMarkTanabe}, a pre-specified number of solutions should be selected from non-dominated solutions in the UEA as the final solution set. In this scenario, distance-based subset selection \cite{DSS, DSSp}, hypervolume-based subset selection \cite{HypE} and IGD-based subset selection \cite{HisaoIJCAI} are usually employed.

In recent years, some EMO algorithms with clustering-based subset selection, such as CLUMOEA \cite{CLUMOEA} and CA-MOEA \cite{CA-MOEA}, achieve the state-of-the-art performance on a variety of multiobjective optimization problems. Besides, in some adaptive reference vector-based EMO algorithms, clustering algorithms are used to update reference vectors (e.g., SRV \cite{ClusterRV}). In the UEA scenario, Singh et al. \cite{DSS} use K-means clustering as a plausible baseline for distance-based subset selection. Although clustering-based subset selection is adopted in many scenarios, different methods have never been analyzed and compared in detail in the literature.

In the past few decades, a huge number of clustering algorithms have been proposed and examined in the literature. However, they are not proposed specially for subset selection in EMO algorithms. Thus, the objective of clustering and the basic assumption about data distributions are different between its traditional applications and its use in EMO algorithms. Currently, only a few clustering algorithms have been used for subset selection in EMO algorithms, which are not always the optimal choice. How to select a clustering algorithm in different EMO scenarios is a difficult problem.

To bridge the gap between clustering algorithms and subset selection in EMO algorithms, in this paper, the performance of different clustering approaches for subset selection on various data sets is examined and analyzed. The main contributions of our paper can be summarized as follows:
\begin{enumerate}
    \item We point out that IGD-based subset selection can be viewed as clustering-based subset selection.
    \item Various promising clustering algorithms are examined on representative data sets. Based on experimental results, we derive some suggestions about the choice of clustering algorithms in different scenarios.
    \item We demonstrate that clustering-based subset selection can be extended to include decision-maker's preference. 
\end{enumerate}

The remainder of this paper is organized as follows. We first review clustering algorithms and IGD-based subset selection, and discuss their relation in Section II. We also discuss the difference between general applications of clustering and clustering-based subset selection in Section II. Then, through computational experiments, we compare and analyze the performance of different representative point selection methods and different clustering algorithms in Section III. Based on experimental results, some suggestions about the choice of clustering methods are given. In Section IV, we further extend clustering-based subset selection to include the decision-maker's preference. Finally, we draw a brief conclusion and suggest some feature research directions in Section V. 

\section{Clustering}
\subsection{General Clustering Algorithms}
For a given data set $D = \{\bm{x}_1, \bm{x}_2,..., \bm{x}_n\}$, a clustering algorithm divides $D$ into $k$ clusters $C = \{C_1, C_2, ..., C_k\}$ where $\forall i \ne j, C_i \cap C_j = \emptyset$ and $D = \cup_{i=1}^k C_i$.

Partition-based clustering algorithms, such as K-means \cite{kmeans}, K-means++ \cite{km++}, and K-medoids \cite{kmedoids}, are the most widely-used clustering algorithms. In this type of clustering algorithms, $k$ cluster centers are first initialized randomly as in K-means or based on some heuristics as in K-means++. Then, each data point is assigned to the closest cluster center. The data points assigned to the same cluster center belong to the same cluster. After that, the cluster center is updated. In K-means and K-means++, the center of each cluster is updated to the mean (centroid) of all data points in that cluster. In K-medoids, the data point with the smallest total distance to all the other points in each cluster is chosen as the new cluster center of the corresponding cluster. Therefore, the objective of K-medoids clustering is
\begin{equation}
    \text{Minimize} \sum_{i=1}^k{\sum_{\bm{x}\in C_i}{\left \| \bm{x}-\bm{\mu}_i \right \|}},
\end{equation}
where $\bm{\mu_i} \in C_i$. 

The main disadvantage of partition-based clustering is that it is not suitable for non-convex clusters. In order to deal with clusters of arbitrary shapes, density-based methods (e.g., DBSCAN \cite{DBSCAN}) are proposed. This type of clustering algorithms assume that clusters in high-density areas are separated by low-density areas in the data space. These algorithms can obtain clusters with arbitrary shapes efficiently but clustering results are usually sensitive to the parameter setting.

Another popular approach is hierarchical clustering \cite{HC}, which aims to construct a hierarchical relationship among the given points. The agglomerative approach is usually employed in hierarchical clustering. Each point is considered as a separate cluster at the initial stage. Then, in each iteration, two clusters that are closest to each other are found and merged. There are various criteria (i.e., linkage functions) to evaluate the distance between two clusters, such as single linkage, complete linkage, weighted average linkage \cite{weighted} and Ward's linkage \cite{ward}. Single linkage calculates the smallest distance between data points in two clusters while complete linkage calculates the largest distance between them. Weighted average linkage calculates the distance between two clusters in a recursive way. If cluster $r$ is obtained by merging clusters $p$ and $q$, the distance between clusters $r$ and $s$ is calculate by $d(r, s) = \frac{d(p, s) + d(q, s)}{2}$.
The Ward's linkage calculates the incremental sum of squares, which can be written as $d(r, s) = \sqrt{\frac{2n_r n_s}{n_r + n_s}}\left \| \bm{\mu}_r-\bm{\mu}_s \right \|,
$
where $n_r$ and $n_s$ are the number of data points in clusters $r$ and $s$, respectively, and $\bm{\mu}_r$ and $\bm{\mu}_s$ are the cluster centers of clusters $r$ and $s$, respectively.

Besides, there are many other types of clustering algorithms that aim to deal with data sets with various characteristics, such as model-based clustering \cite{GMM}, grid-based clustering \cite{sting}, fuzzy clustering \cite{FCM}. For more details about clustering algorithms, please refer to \cite{survey}.  

\subsection{Clustering-Based Subset Selection}
Since clustering algorithms can divide a data set into a pre-specified number of clusters, clustering algorithms can be used for subset selection in EMO algorithms \cite{CLUMOEA,CA-MOEA,DSS}. To select a subset of size $k$ from a given data set, we first apply a clustering algorithm to the data set. Based on the obtained $k$ clusters, a representative point will be selected from each cluster. These representative points compose the selected subset. 

There are some differences between the general use of clustering algorithms and the use of clustering algorithms for subset selection in EMO algorithms. Firstly, since the subset size is usually pre-specified, clustering-based subset selection prefers algorithms that can obtain a fixed number of clusters instead of a variable number of clusters. Besides, some clustering algorithms are designed to deal with complicated cluster shapes or special distributions, which is of little use for subset selection. Additionally, the pre-specified number of clusters is usually much larger in clustering-based subset selection than that in standard applications of clustering algorithms.

Due to these differences, clustering algorithms that perform well in their standard applications may not be suitable for clustering-based subset selection. The performance of clustering algorithms for subset selection are examined in some applications (e.g., NIR spectra analysis \cite{NIR} and IT service \cite{IT}). In our paper, we focus on subset selection in EMO algorithms. More promising clustering algorithms are considered, and different methods for selecting a representative point in each cluster is discussed. To the best of our knowledge, this issue has not been examined in the literature.

\subsection{IGD-based Subset Selection}
Inverted generational distance (IGD) \cite{IGD} is one of the most widely-used indicators to evaluate the performance of an EMO algorithm. It calculates the average distance from each point $r$ in the reference set $R$ to the closest point in the obtained solution set $P$. The IGD contribution of a solution $\bm{p} \in P$ to the solution set $P$ is defined as $ IGDC(\bm{p}, R) = IGD(P\setminus \bm{p}, R) - IGD(P, R).$

Since the IGD indicator considers the convergence and diversity simultaneously, IGD-based subset selection can be used to select representative solutions from an archive. In this case, all solutions in the archive are used as the reference set for IGD calculation. It aims to select a subset $S$ of size $k$ with the smallest IGD value with respect to all solutions in the archive. Formally, the objective of IGD-based subset selection is
\begin{equation}
\text{Minimize  } \frac{1}{n}\sum_{j=1}^n \min_{\bm{s} \in S}{\left \| \bm{x}_j-\bm{s} \right \|},  
\end{equation}
where $\bm{x}_j$ is a point in the objective space included in the archive, and $n$ is the number of points in the archive (i.e., $\{\bm{x}_1, \bm{x}_2, ..., \bm{x}_n\}$ is the reference point set for IGD calculation). 
We can divide the whole data set into $k$ groups $G = \{G_1, G_2, ..., G_k\}$ according to the closest point in the subset $S$ (where $|S| = k$). Let us define an indicator function as:
\begin{equation}
I(i, j) = \left\{
\begin{aligned}
1 & &if \  \bm{x_j} \in G_i,\\
0 & &otherwise. \\
\end{aligned}
\right.
\end{equation} 
Then, the objective of IGD-based subset selection can be re-written as 
\begin{multline}
    \text{Minimize  }  \frac{1}{n}\sum_{j=1}^n \sum_{i=1}^k \left \| \bm{x}_j-\bm{s}_i \right \| I(i, j) \\
= \frac{1}{n}\sum_{i=1}^k \sum_{j=1}^n \left \| \bm{x}_j-\bm{s}_i \right \| I(i, j)  
= \frac{1}{n}\sum_{i=1}^k {\sum_{\bm{x}\in G_i}{\left \| \bm{x}-\bm{s}_i \right \|}},
\end{multline}
which is exactly the same as the objective of K-medoids clustering (i.e., Eq. (1)). Therefore, although IGD-based subset selection is usually classified as indicator-based subset selection, it can also be viewed as a clustering-based subset selection method.

Greedy inclusion is usually used in IGD-based subset selection from large data sets. A greedy inclusion algorithm starts from an empty subset. In each iteration, it selects the data point with the largest IGD contribution to the currently selected subset until a pre-specified number of data points are selected. On the contrary, a greedy removal algorithm starts with the whole data set and iteratively removes the solution with the smallest IGD contribution to the current subset. When the data set size is much larger than the subset size, the greedy removal strategy is not efficient.


 Greedy inclusion IGD-based subset selection can be viewed as a top-down clustering approach. In the beginning, the entire data set is treated as a single cluster. Then, new cluster centers are added one by one until the pre-specified number of clusters is found. For example, Fig. 1 shows a data set with seven data points. Let us assume that we want to select three data points. In the initial stage, all the data points belong to a big cluster whose center is $D$. In the next iteration, the first cluster center $D$ is fixed, and the algorithm aims to find another point to minimize the objective in (4). The cluster center $B$ is found and fixed. Now the data set is divided into two clusters (shaded in green). In the same manner, the third cluster center $F$ is found and the data set is divided into three clusters (shown by three pink dashed circles). 
 
\begin{figure}[htbp]
\centering
\includegraphics[width=0.23 \textwidth, trim=0 0 0 0,clip]{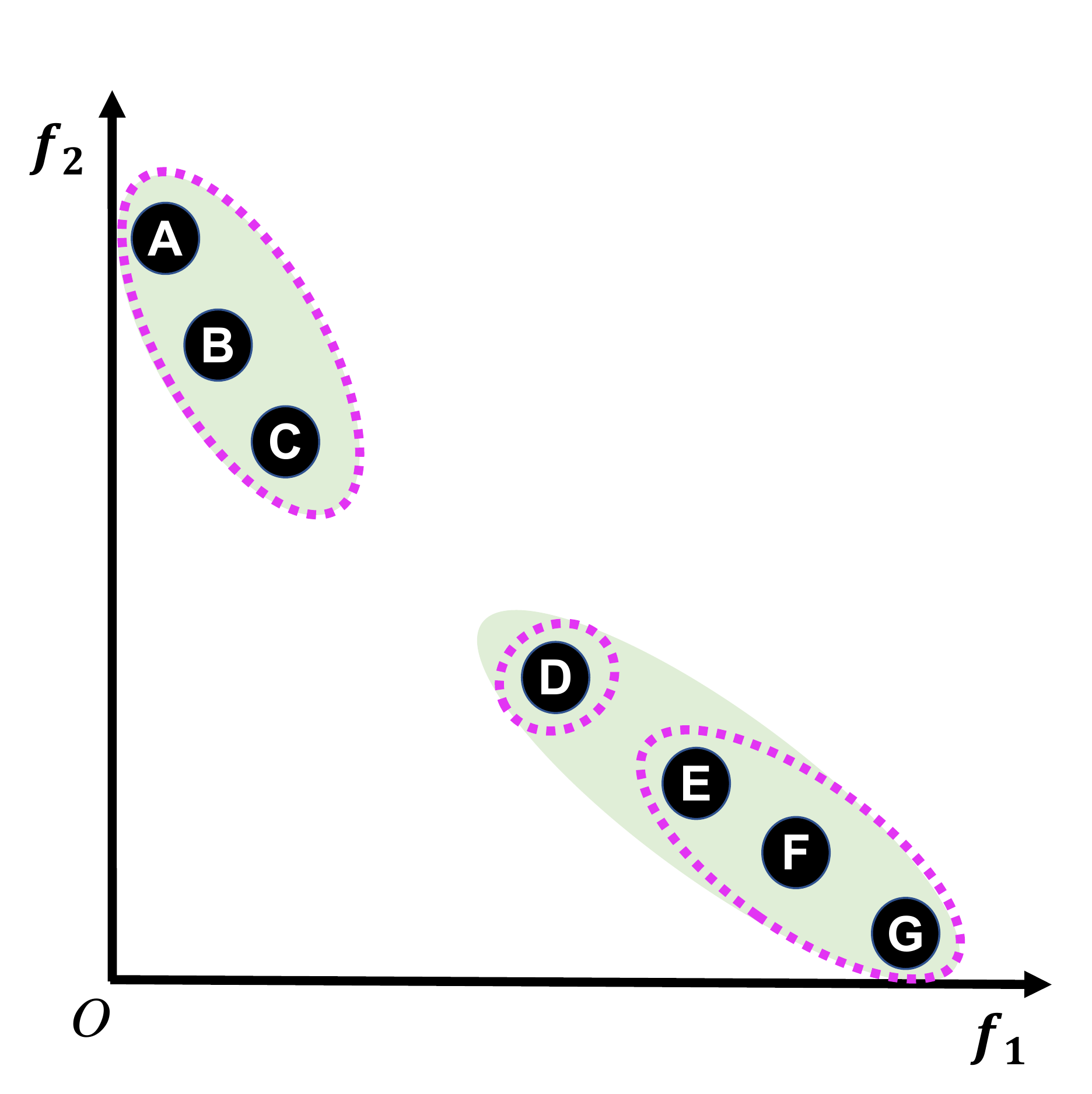}
\caption{Illustration of the greedy inclusion IGD-based subset selection.}
\end{figure}

\section{Experimental Study}
\subsection{Data Sets Generation}
In our experimental study, as a candidate solution set $D$, we consider randomly sampled data points on the true Pareto fronts of six test problems with various Pareto front shapes, including linear (DTLZ1 \cite{DTLZ}), concave (DTLZ2 \cite{DTLZ}), convex (MaF3 \cite{MaF}), inverted (IDTLZ2 \cite{NSGAIII2}), and disconnected (DTLZ7 \cite{DTLZ} and WFG2 \cite{WFG}). 

Three candidate solution sets (i.e., $D$) of different size are generated for each test problem as follows. We first generate a large number of uniformly distributed points on the true Pareto front by the method in Tian et al. \cite{Sampling}. Note that this method cannot generate an arbitrary number of points. In our experiments, the number of generated points is specified as the number closest to 200,000 among the number of points that this method can generate. Then, we randomly select 200, 4,000, and 10,000 points from these uniformly generated points. The data set of size 200 is used to test the performance in the environmental selection scenario while the data sets of size 4,000 and 10,000 are used to test the performance in the final solution selection from the UEA scenario.

\subsection{Compared Algorithms}
We examine six promising and well-known clustering algorithms on each candidate data set. The details of the examined algorithms are as follows. 
\begin{enumerate}
    \item K-means++. An improved version of K-means with heuristic cluster center initialization. The maximum number of iterations is specified as 100. 
    \item K-medoids. We use the same cluster center initialization method as K-means++. The maximum number of iterations is specified as 100.
    \item Greedy IGD-based subset selection (Greedy IGD). Since the select subset size $k$ is smaller than or equal to half of the data set size $n$, we choose greedy inclusion IGD-based subset selection. Lazy greedy inclusion IGD-based subset selection \cite{LGI} is used to accelerate the selection. The obtained subset by the lazy greedy inclusion algorithm is exactly the same as the naive greedy inclusion algorithm. 
    \item Three variants of of Hierarchical Clustering (HC). Following common practice, we use the agglomerative hierarchical clustering approach. Three linkage functions are tested, i.e., Ward's linkage, weighted average linkage and single linkage. 
\end{enumerate}

Since density-based clustering algorithms and grid-based clustering algorithms cannot generate a pre-specified number of clusters, they are not examined in this paper. Model-based clustering algorithms are very expensive when the data set is large. Therefore, they are not suitable for our test scenarios.

In all experiments in this section, the number of clusters (i.e., subset size) is set to 100. The IGD value of the obtained subset with respect to the corresponding candidate data set (i.e., with respect to the reference point set for the IGD calculation) is used to evaluate the performance of each algorithm. All the experiments are carried out on the server equipped with Intel Xeon E5-2620.

\subsection{How to Select a Representative Point in Each Cluster?}
In some clustering algorithms, such as K-medoids and IGD-based subset selection, the representative point of each cluster is obtained as the cluster center when the execution of the clustering algorithm is terminated. We can directly use these points (i.e., the obtained cluster centers) as the final subset. However, in K-means++, although a pre-specified number of cluster centers are determined, each cluster center is not a data point but the mean (centroid) of the data points in the cluster. In hierarchical clustering algorithms, no clear cluster center is determined. Therefore, we need to select a representative point for each cluster in those clustering algorithms where the cluster center is not a data point or no cluster center is explicitly calculated. We can choose the representative point using the following two strategies.
\begin{enumerate}
    \item Strategy 1: Choose the data point closest to the mean (centroid) of the data points in the cluster.
    \begin{equation}
        r_i = {argmin}_{\bm{x} \in C_i} \left \| \bm{x}-\bm{\bar{x}} \right \|
    \end{equation}
    \item Strategy 2: Choose the point with the smallest total distance to all data points in the cluster. 
    \begin{equation}
        r_i = {argmin}_{\bm{x} \in C_i} \sum_{\bm{y} \in C_i} \left \| \bm{y}-\bm{x} \right \|
    \end{equation}
\end{enumerate}

Table I shows the comparison results between these two strategies on the clusters obtained by K-means++ and hierarchical clustering with Ward's linkage. We compare these two strategies using the Wilcoxon rank-sum test \cite{rank_sum_test} with the significance level 0.1. The "+" and "-" in the table mean Strategy 2 is significantly better and significantly worse than Strategy 1, respectively, while "=" means there is no statistically significant difference between the two strategies.

As shown in Table I, we can see that when the data set size is 200, there is no significant difference between these two strategies. The reason is that the expected number of data points in each cluster is two when the number of clusters $k$ is set to 100. For a cluster with two data points, if Strategy 1 is used, the cluster center $\bm{\bar{x}}$ is the midpoint of the two data points, and the distance from $\bm{\bar{x}}$ to each data point is the same. Therefore, one of the two points is randomly selected as the representative point. If Strategy 2 is used, the distance from one point to the other point is the same (independent of the choice of one of the two points). Thus, one of the two points is randomly selected as the representative point. Thus, the difference between these two strategies is not significant.

When the size of the data set is larger, we can see that Strategy 2 always has the better or equal performance to Strategy 1 since the objective of Strategy 2 is consistent with IGD minimization while the objective of Strategy 1 is slightly different from IGD minimization. Besides, Strategy 2 is more robust to outliers in the cluster. Therefore, we use Strategy 2 to select representative data points in the remaining experiments. 

\begin{table}[htbp]
\centering
\caption{Comparison of Strategy 2 with Strategy 1 for representative point selection on the clustering results by K-Means++ and hierarchical clustering for the six test problems. "+" means that Strategy 2 obtains better IGD than Strategy 1.}
\label{tab:my-table}
\setlength{\tabcolsep}{6mm}{
\begin{tabular}{@{}cccc@{}}
\toprule
& Data Set & K-Means++ & HC (Ward) \\
& Size    & +/=/-     & +/=/-     \\ \midrule
\multirow{3}{*}{3-objective} & 200     & 0/6/0     & 0/6/0     \\
& 4000    & 1/5/0     & 6/0/0     \\
& 10000   & 3/3/0     & 6/0/0     \\ \midrule
\multirow{3}{*}{8-objective} & 200     & 0/6/0     & 0/6/0     \\& 4000    & 3/3/0     & 5/1/0     \\
& 10000   & 2/4/0     & 4/2/0     \\ \bottomrule
\end{tabular}}
\end{table}

\subsection{Performance on Three-Objective Data Sets}
Table II shows the average IGD value of the subsets obtained by each algorithm on each three-objective data set over 11 runs. The number in parentheses shows the ranking of the corresponding algorithm (among the six algorithms) for the corresponding test problem.

\begin{table*}[htbp]
\footnotesize
\centering
\caption{Average IGD value of the subsets obtained by each algorithm on each three-objective data set over 11 runs}
\label{tab:my-table}
\begin{tabular}{@{}llllllll@{}}
\toprule
\textbf{Data set size}& \textbf{Shape}     & \textbf{K-means++} & \textbf{K-medoids} & \textbf{Greedy IGD} & \textbf{HC (ward)} & \textbf{HC (weighted)} & \textbf{HC (single)} \\ \midrule
\multirow{7}{*}{\textbf{\begin{tabular}[c]{@{}l@{}}200 \end{tabular}}}   & Linear             & 1.5605E-02 (6)     & 1.5046E-02 (4)     & 1.4034E-02 (3)      & 1.3951E-02 (2)   & 1.3815E-02 (1)     & 1.5251E-02 (5)     \\
& Concave            & 1.9853E-02 (6)     & 1.9256E-02 (4)     & 1.8238E-02 (3)      & 1.8126E-02 (2)   & 1.7905E-02 (1)     & 1.9356E-02 (5)     \\
& Convex             & 1.1120E-02 (6)     & 1.0633E-02 (4)     & 9.9369E-03 (2)      & 9.8542E-03 (1)   & 9.9591E-03 (3)     & 1.0866E-02 (5)     \\
& Inverted           & 2.1808E-02 (6)     & 2.1619E-02 (5)     & 2.0249E-02 (3)      & 1.9523E-02 (2)   & 1.9387E-02 (1)     & 2.1053E-02 (4)     \\
& Disconnected 1      & 2.0288E-02 (5)     & 1.9955E-02 (4)     & 1.8635E-02 (3)      & 1.8026E-02 (1)   & 1.8094E-02 (2)     & 2.2321E-02 (6)     \\
& Disconnected 2      & 5.1448E-02 (5)     & 5.0210E-02 (4)     & 4.6697E-02 (3)      & 4.5358E-02 (2)   & 4.5127E-02 (1)     & 5.3329E-02 (6)     \\
& \textbf{Avg. Rank} & \textbf{5.67}      & \textbf{4.16}      & \textbf{2.83}       & \textbf{1.67}    & \textbf{1.5}       & \textbf{5.17}      \\ \midrule
\multirow{7}{*}{\textbf{\begin{tabular}[c]{@{}l@{}}4000 \end{tabular}}}  & Linear             & 3.4245E-02 (1)     & 3.4610E-02 (3)     & 3.5086E-02 (4)      & 3.4267E-02 (2)   & 3.5127E-02 (5)     & 5.4144E-02 (6)     \\
& Concave            & 4.5270E-02 (1)     & 4.5841E-02 (3)     & 4.6600E-02 (4)      & 4.5425E-02 (2)   & 4.6955E-02 (5)     & 9.7713E-02 (6)     \\
& Convex             & 2.4799E-02 (2)     & 2.5095E-02 (3)     & 2.5222E-02 (4)      & 2.4690E-02 (1)   & 2.7967E-02 (5)     & 1.1114E-01 (6)     \\
& Inverted           & 4.5342E-02 (2)     & 4.5933E-02 (3)     & 4.6540E-02 (4)      & 4.5275E-02 (1)   & 4.7141E-02 (5)     & 1.3222E-01 (6)     \\
& Disconnected 1      & 4.2280E-02 (2)     & 4.2566E-02 (3)     & 4.3181E-02 (4)      & 4.1740E-02 (1)   & 4.4241E-02 (5)     & 1.1989E-01 (6)     \\
& Disconnected 2      & 1.1831E-01 (2)     & 1.1997E-01 (4)     & 1.1970E-01 (3)      & 1.1717E-01 (1)   & 1.3169E-01 (5)     & 4.2332E-01 (6)     \\
& \textbf{Avg. Rank} & \textbf{1.67}      & \textbf{3.16}      & \textbf{3.83}       & \textbf{1.33}    & \textbf{5}         & \textbf{6}         \\ \midrule
\multirow{7}{*}{\textbf{\begin{tabular}[c]{@{}l@{}}10000 \end{tabular}}} & Linear             & 3.5000E-02 (1)     & 3.5515E-02 (3)     & 3.6214E-02 (4)      & 3.5495E-02 (2)   & 3.6623E-02 (5)     & 5.7334E-02 (6)     \\
& Concave            & 4.6263E-02 (1)     & 4.6975E-02 (3)     & 4.7350E-02 (4)      & 4.6844E-02 (2)   & 4.8787E-02 (5)     & 1.3004E-01 (6)     \\
& Convex             & 2.5000E-02 (1)     & 2.5479E-02 (3)     & 2.5527E-02 (4)      & 2.5258E-02 (2)   & 2.8881E-02 (5)     & 1.2888E-01 (6)     \\
& Inverted           & 4.6183E-02 (1)     & 4.6816E-02 (2)     & 4.7502E-02 (4)      & 4.6865E-02 (3)   & 4.8578E-02 (5)     & 1.3600E-01 (6)     \\
& Disconnected 1      & 4.2736E-02 (1)     & 4.3048E-02 (3)     & 4.4097E-02 (4)      & 4.2901E-02 (2)   & 4.5914E-02 (5)     & 1.1568E-01 (6)     \\
& Disconnected 2      & 1.2030E-01 (1)     & 1.2231E-01 (3)     & 1.2324E-01 (4)      & 1.2134E-01 (2)   & 1.3523E-01 (5)     & 4.6655E-01 (6)     \\
& \textbf{Avg. Rank} & \textbf{1}         & \textbf{2.83}      & \textbf{4}          & \textbf{2.17}    & \textbf{5}         & \textbf{6}         \\ \bottomrule
\end{tabular}
\end{table*}

When the data set size is 200, hierarchical clustering with weighted average linkage achieves the best average rank while K-means++ gets the worst average rank. However, when the data set size increases, which is similar to subset selection from the archive, the performance of K-means++ clearly improves and the performance of hierarchical clustering with weighted average linkage clearly degrades. When the data set size is 10,000, K-means++ always achieves the best performance. This shows that K-means++ is not suitable when the number of data points in a cluster is too small. This is because the distance between the cluster center of K-means++ and the selected representative point is large when the data set size is small. By increasing the data set size, this distance becomes small. As a result, the performance of K-means++ as a subset selection method is improved. For higher-dimensional problems (e.g., Table III on the 8-objective data sets), much more data points are needed to have a close data point to each cluster center. Thus, K-means++ is not the best even in the case of 10,000 data points.  

\begin{figure}[htbp]
\centering
  \subfigure[200 data points.] {\includegraphics[width=0.30\textwidth,trim=0 0 0 0,clip]{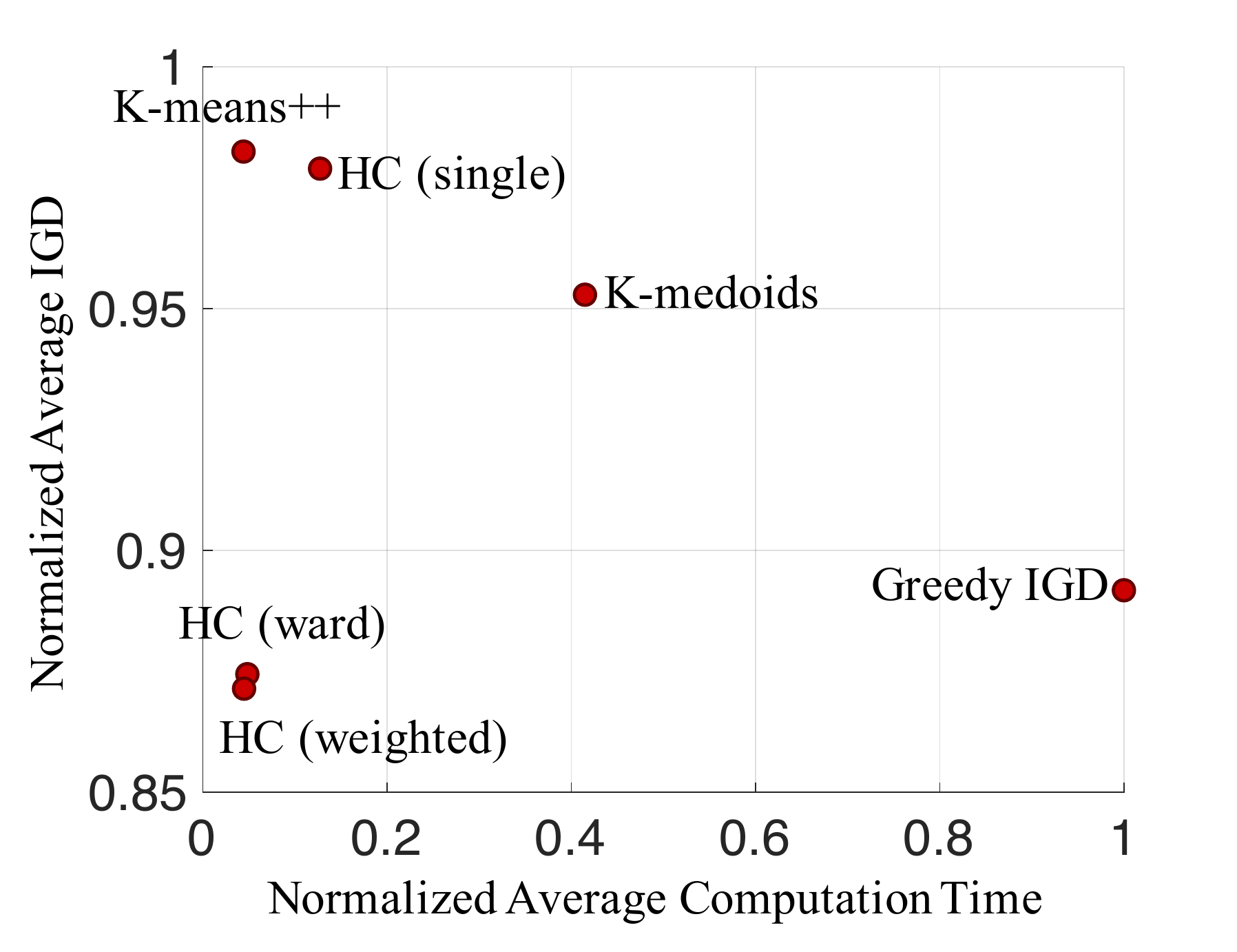}}
  \subfigure[10,000 data points.] {\includegraphics[width=0.30\textwidth,trim=0 0 0 0,clip]{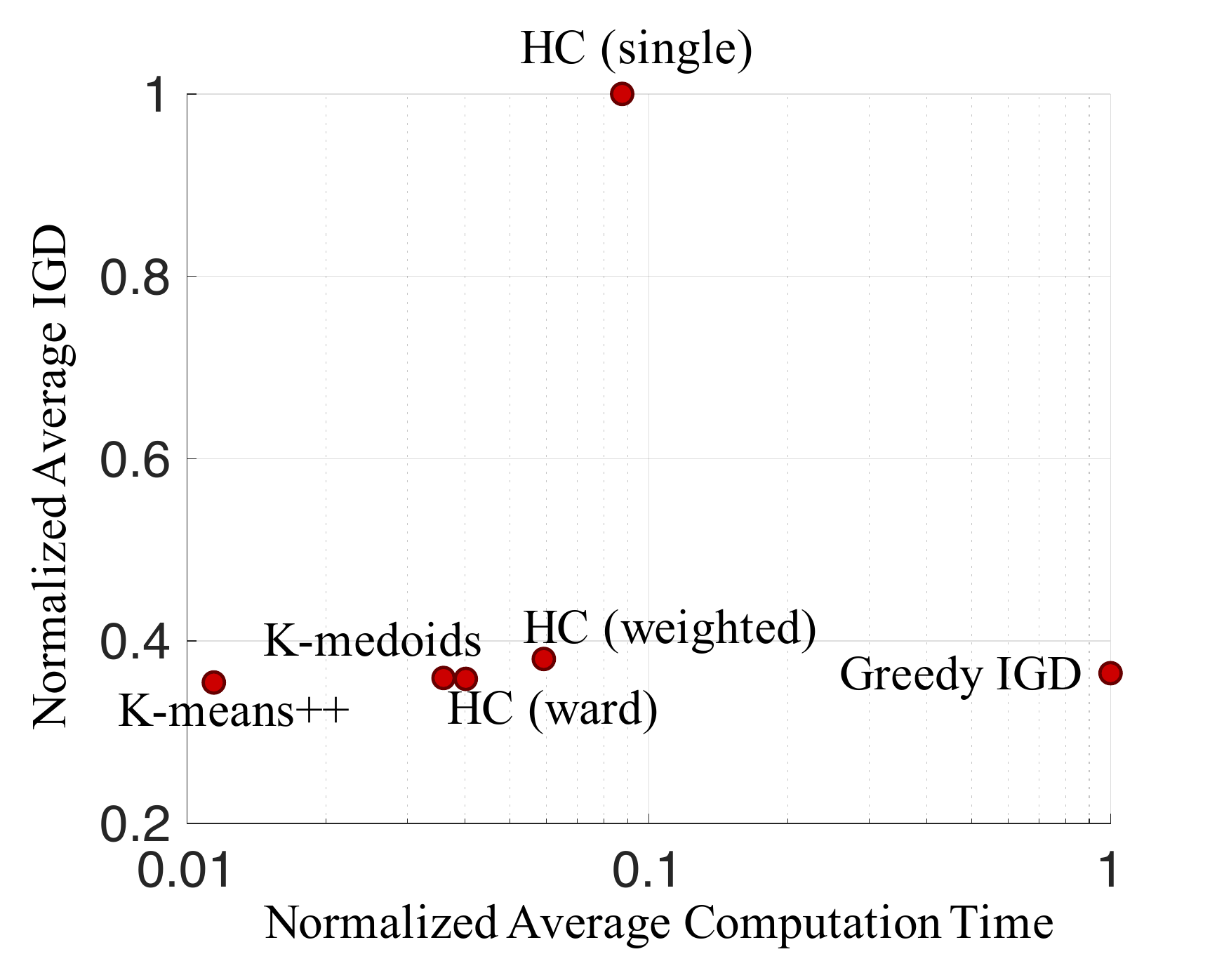}}
  \caption{Relation between the average IGD value and the average computation time of each algorithm on the three-objective data sets (each axis is normalized using the largest value).}
\end{figure}

Computation time is also important in evaluating clustering algorithms. Fig. 2 shows the relation between the average IGD value of the obtained subsets and the average computation time of each algorithm for different data set sizes. Since the IGD values of the selected subsets from data sets on different Pareto fronts have large differences (depending on the Pareto front shape), each IGD value is normalized by dividing the largest IGD value obtained for each data set before calculating the average value. To be consistent, we do the same normalization on the computation time. On the data sets of size 10,000, since the computation time of greedy IGD-based subset selection is much larger than that of the other algorithms, we use the log scale when plotting the figure.

As shown in Fig. 2(a), when the data set size is small (i.e., 200), hierarchical clustering with weighted average linkage dominates all the other algorithms (with respect to both the IGD value and the computation time), and hierarchical clustering with Ward's linkage achieves similar performance. When the data set size is large (i.e., 10,000), we can observe from Fig. 2(b) that K-means++ is the most efficient one and achieves the best average IGD value. K-medoids and hierarchical clustering with Ward's linkage are similar in terms of the IGD value and the computation time.

In general, on the three-objective data sets, if the data set size is small, hierarchical clustering with weighted average linkage is the best choice among the examined algorithms. If the data set size is large, K-means++, which can obtain the best IGD value in the shorted computation time is the best choice.

\begin{table*}[htbp]
\footnotesize
\centering
\caption{Average IGD value of the subsets obtained by each algorithm on each eight-objective data set over 11 runs}
\label{tab:my-table}
\begin{tabular}{@{}llllllll@{}}
\toprule
\textbf{Data set size}& \textbf{Shape}     & \textbf{K-means++} & \textbf{K-medoids} & \textbf{Greedy IGD} & \textbf{HC ward} & \textbf{HC (weight)} & \textbf{HC (single)} \\ \midrule
\multirow{7}{*}{\textbf{\begin{tabular}[c]{@{}l@{}}200 \end{tabular}}}   & Linear             & 9.2909E-02 (5)     & 8.9396E-02 (4)     & 8.6373E-02 (1)      & 8.7361E-02 (3)   & 8.7329E-02 (2)     & 9.8237E-02 (6)     \\
& Concave            & 1.7983E-01 (6)     & 1.7308E-01 (4)     & 1.6833E-01 (1)      & 1.6986E-01 (2)   & 1.6999E-01 (3)     & 1.7969E-01 (5)     \\
& Convex             & 2.5646E-02 (5)     & 2.4751E-02 (4)     & 2.3729E-02 (1)      & 2.4097E-02 (2)   & 2.4464E-02 (3)     & 2.9236E-02 (6)     \\
& Inverted           & 1.7442E-01 (5)     & 1.6856E-01 (4)     & 1.6219E-01 (1)      & 1.6429E-01 (3)   & 1.6349E-01 (2)     & 1.7501E-01 (6)     \\
& Disconnected 1      & 3.0614E-01 (5)     & 2.9561E-01 (4)     & 2.8529E-01 (1)      & 2.9072E-01 (3)   & 2.8885E-01 (2)     & 3.1355E-01 (6)     \\
& Disconnected 2      & 2.7587E-01 (5)     & 2.6896E-01 (4)     & 2.5453E-01 (1)      & 2.5721E-01 (3)   & 2.5477E-01 (2)     & 3.5119E-01 (6)     \\
& \textbf{Avg. Rank} & \textbf{5.17}      & \textbf{4}         & \textbf{1}          & \textbf{2.67}    & \textbf{2.33}      & \textbf{5.83}      \\ \midrule
\multirow{7}{*}{\textbf{\begin{tabular}[c]{@{}l@{}}4000 \end{tabular}}}  & Linear             & 1.7655E-01 (2)     & 1.7679E-01 (3)     & 1.7381E-01 (1)      & 1.7755E-01 (4)   & 1.8155E-01 (5)     & 2.3244E-01 (6)     \\
& Concave            & 3.3520E-01 (2)     & 3.3587E-01 (3)     & 3.3000E-01 (1)      & 3.3711E-01 (4)   & 3.4313E-01 (5)     & 4.4839E-01 (6)     \\
& Convex             & 5.4729E-02 (2)     & 5.5212E-02 (4)     & 5.3800E-02 (1)      & 5.4770E-02 (3)   & 6.2500E-02 (5)     & 9.2508E-02 (6)     \\
& Inverted           & 3.3606E-01 (3)     & 3.3589E-01 (2)     & 3.3106E-01 (1)      & 3.3721E-01 (4)   & 3.4376E-01 (5)     & 4.6123E-01 (6)     \\
& Disconnected 1      & 5.6078E-01 (3)     & 5.5921E-01 (2)     & 5.5612E-01 (1)      & 5.6186E-01 (4)   & 5.8916E-01 (5)     & 7.5353E-01 (6)     \\
& Disconnected 2      & 6.2260E-01 (2)     & 6.2575E-01 (4)     & 6.1100E-01 (1)      & 6.2471E-01 (3)   & 7.1781E-01 (5)     & 1.1343E+00 (6)     \\
& \textbf{Avg. Rank} & \textbf{2.33}      & \textbf{3}         & \textbf{1}          & \textbf{3.67}    & \textbf{5}         & \textbf{6}         \\ \midrule
\multirow{7}{*}{\textbf{\begin{tabular}[c]{@{}l@{}}10000 \end{tabular}}} & Linear             & 1.7777E-01 (2)     & 1.7977E-01 (3)     & 1.7637E-01 (1)      & 1.7994E-01 (4)   & 1.8207E-01 (5)     & 2.1790E-01 (6)     \\
& Concave            & 3.3675E-01 (2)     & 3.4223E-01 (4)     & 3.3569E-01 (1)      & 3.4089E-01 (3)   & 3.4824E-01 (5)     & 4.5376E-01 (6)     \\
& Convex             & 5.5902E-02 (2)     & 5.6630E-02 (4)     & 5.5219E-02 (1)      & 5.6231E-02 (3)   & 6.5573E-02 (5)     & 1.0163E-01 (6)     \\
& Inverted           & 3.3701E-01 (2)     & 3.4166E-01 (3)     & 3.3555E-01 (1)      & 3.4176E-01 (4)   & 3.4809E-01 (5)     & 4.5171E-01 (6)     \\
& Disconnected 1      & 5.7045E-01 (2)     & 5.7501E-01 (3)     & 5.7029E-01 (1)      & 5.7680E-01 (4)   & 5.9937E-01 (5)     & 7.2398E-01 (6)     \\
& Disconnected 2      & 6.3377E-01 (2)     & 6.4113E-01 (4)     & 6.2533E-01 (1)      & 6.3674E-01 (3)   & 7.6525E-01 (5)     & 1.2024E+00 (6)     \\
& \textbf{Avg. Rank} & \textbf{2}         & \textbf{3.5}       & \textbf{1}          & \textbf{3.5}     & \textbf{5}         & \textbf{6}         \\ \bottomrule
\end{tabular}
\end{table*}

\subsection{Performance on Eight-Objective Data Sets}
We use the eight-objective data sets to test the performance of each algorithm in many-objective scenarios. Table III shows the average IGD value of the subsets obtained by each algorithm on each eight-objective
data set over 11 runs. 

Although greedy IGD-based subset selection does not have very good performance in the three-objective case in Table II, it always achieves the best performance in the eight-objective case in Table III. We can see that in the three-objective case (Table II), hierarchical clustering with Ward's linkage consistently outperforms greedy IGD-based subset selection. However, in the eight-objective case, greedy IGD-based subset selection always outperforms hierarchical clustering with Ward's linkage. This is due to the curse of dimensionality. The linkage functions to measure the distance between two clusters are usually designed for clustering in two or three dimensions. Thus, the performance of hierarchical clustering tends to degrade as the number of dimensions increases. Such performance deterioration can also be observed on the other clustering algorithms from the comparison between Table II and Table III. Besides, we can observe in Table III that the rank of hierarchical clustering with weighted average linkage significantly decreases when the data set size increases as in Table II.

\begin{figure}[htbp]
\centering
  \subfigure[200 data points.] {\includegraphics[width=0.30\textwidth,trim=0 0 0 0,clip]{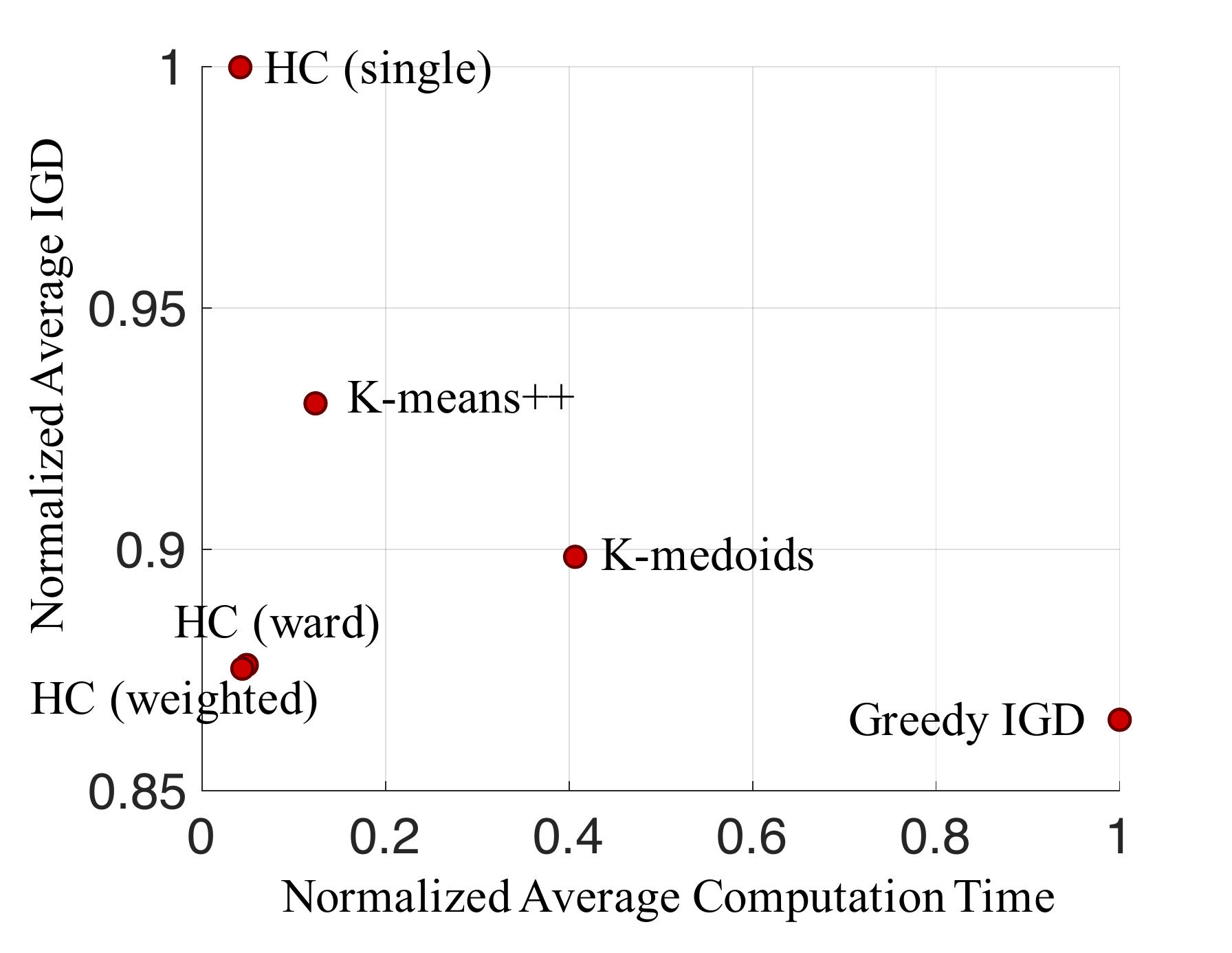}}
  \subfigure[10,000 data points.] {\includegraphics[width=0.30\textwidth,trim=0 0 0 0,clip]{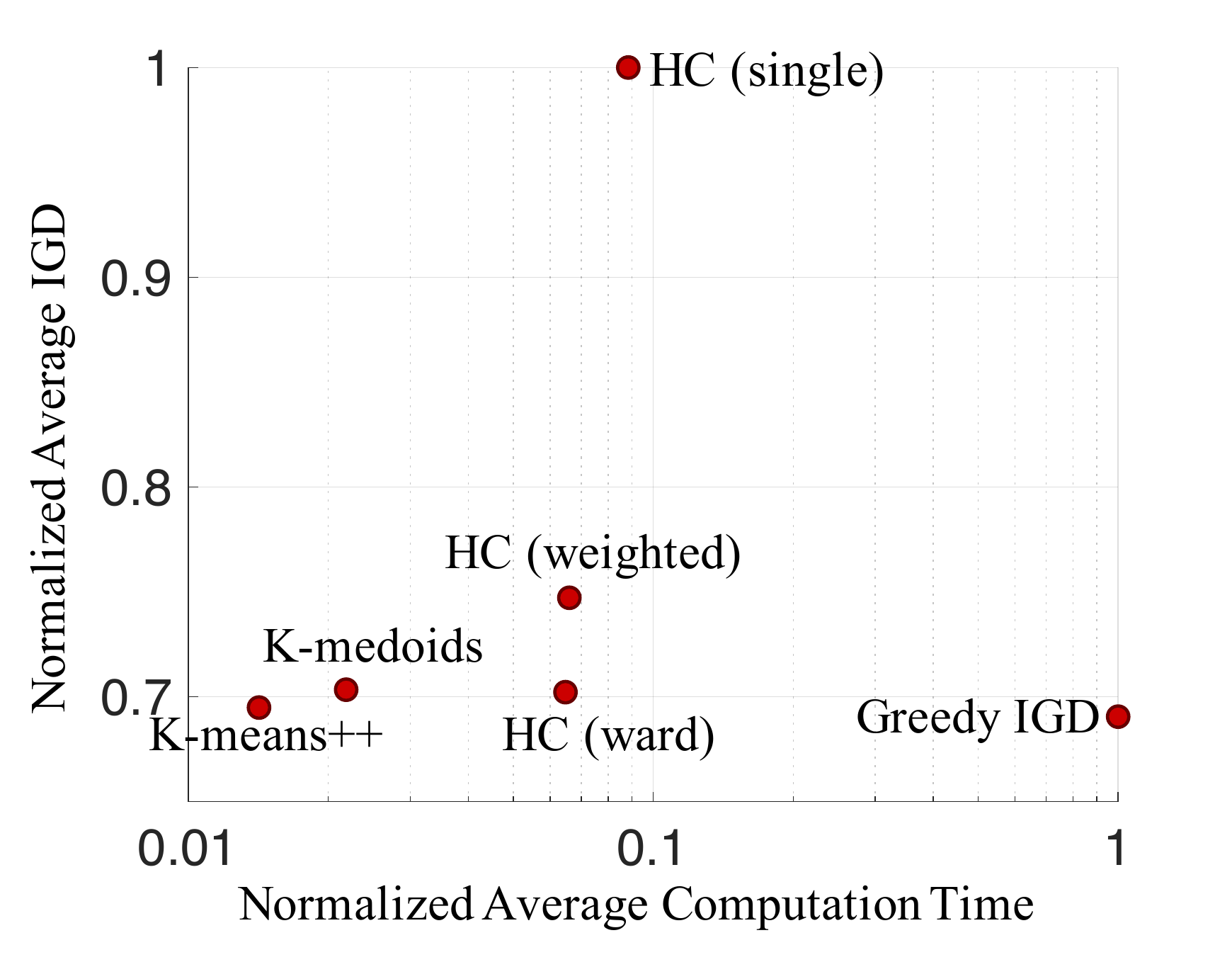}}
  \caption{Relation between the average IGD value and the average computation time of each algorithm on the eight-objective data sets (each axis is normalized using the largest value).}
\end{figure}

In Fig. 3, we show the relation between the average IGD value of the obtained subsets and the average computation time of each algorithm on the eight-objective data sets. The normalization is performed in the same manner as in Fig. 2. 

We can see from Fig. 3(a) that greedy IGD-based subset selection and hierarchical clustering with weighted average linkage are non-dominated to each other. Although greedy IGD-based subset selection can achieve the best IGD performance, it uses a much longer computation time than hierarchical clustering. Hierarchical clustering with weighted average linkage is a cheaper alternative with a slight performance deterioration. In Fig. 3(b), we can see that K-means++ and greedy IGD-based subset selection are non-dominated to each other. All the other algorithms are dominated by K-means++ in Fig. 3(b).

On the eight-objective data sets, greedy IGD-based subset selection always achieves the best IGD performance among the examined algorithms. If the computation time is severely limited, hierarchical clustering with weighted average linkage is a cheaper alternative when the data set size is small, and K-means++ is a cheaper alternative when the data set size is large.

\section{Introduce Preference into Clustering}
Clustering-based subset selection tends to select uniformly distributed data points from a given data set. However, we may prefer some special solutions among the solutions obtained by an EMO algorithm (i.e., prefer some parts of the Pareto front). For example, in some applications, decision-makers are interested in knee regions \cite{Knee2K} since solutions in these regions achieve a better trade-off between objectives. For a solution in the knee region, a small improvement in one objective may cause a large deterioration in other objectives. 

To select more useful solutions for decision-makers, we can introduce the preference for knee regions to the K-medoids clustering algorithm by modifying the medoids update and data points assignment procedure. When updating the medoids, instead of using Euclidean distance, we can use the similarity measure in the IGD+ indicator \cite{IGD+}: $ s(\bm{x}, \bm{y}) = \sqrt{\sum_{i=1}^m{\max(y_i - x_i, 0)^2}}.$ That is, the distance is calculated only for inferior objectives of $\bm{y}$ than $\bm{x}$. Therefore, in each iteration, we aim to find a new medoid $\bm{\mu}_i$ for each cluster which minimizes $\sum_{\bm{x} \in C_i} s(\bm{x}, \bm{\mu}_i)$. Then, each data point $\bm{x}$ is re-assigned to the medoid $\bm{\mu}_l$ with the smallest $s(\bm{x}, \bm{\mu}_l)$.

 \begin{figure*}[htbp]
\centering
  \subfigure[Linear.] {\includegraphics[width=0.42\textwidth,trim=0 0 0 0,clip]{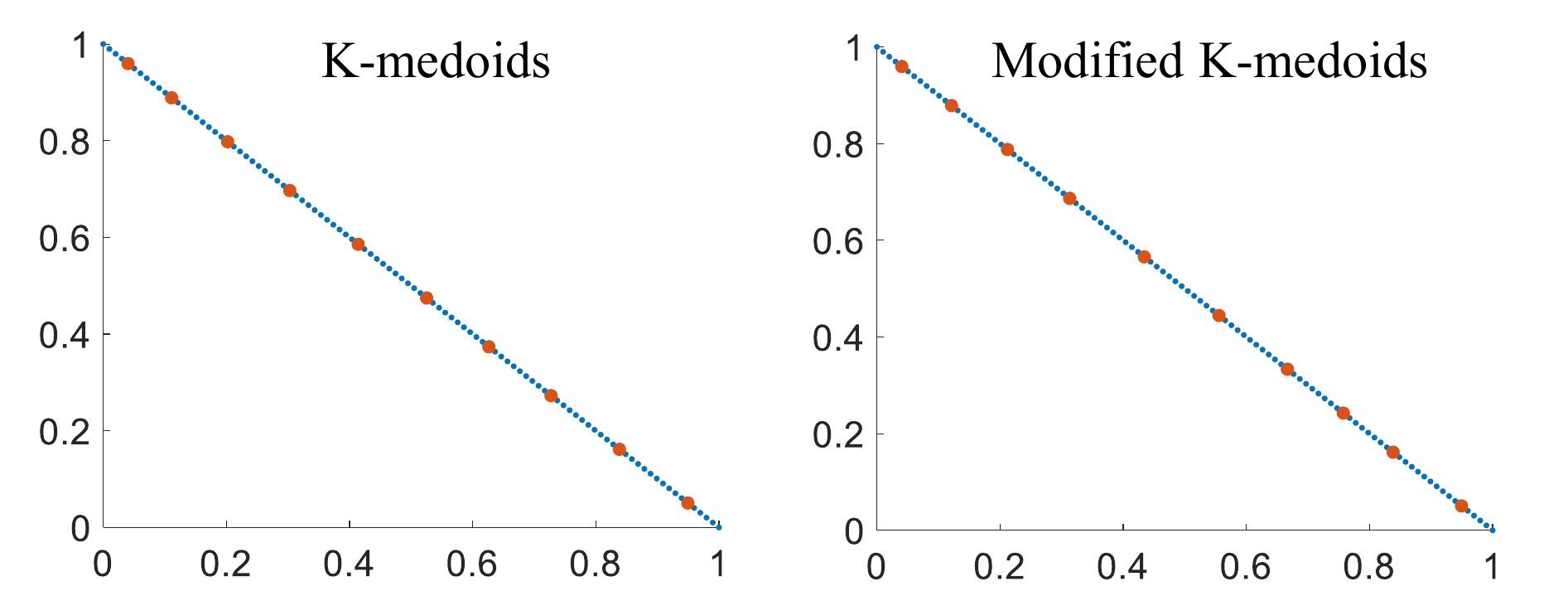}}
  \subfigure[Convex.] {\includegraphics[width=0.42\textwidth,trim=0 0 0 0,clip]{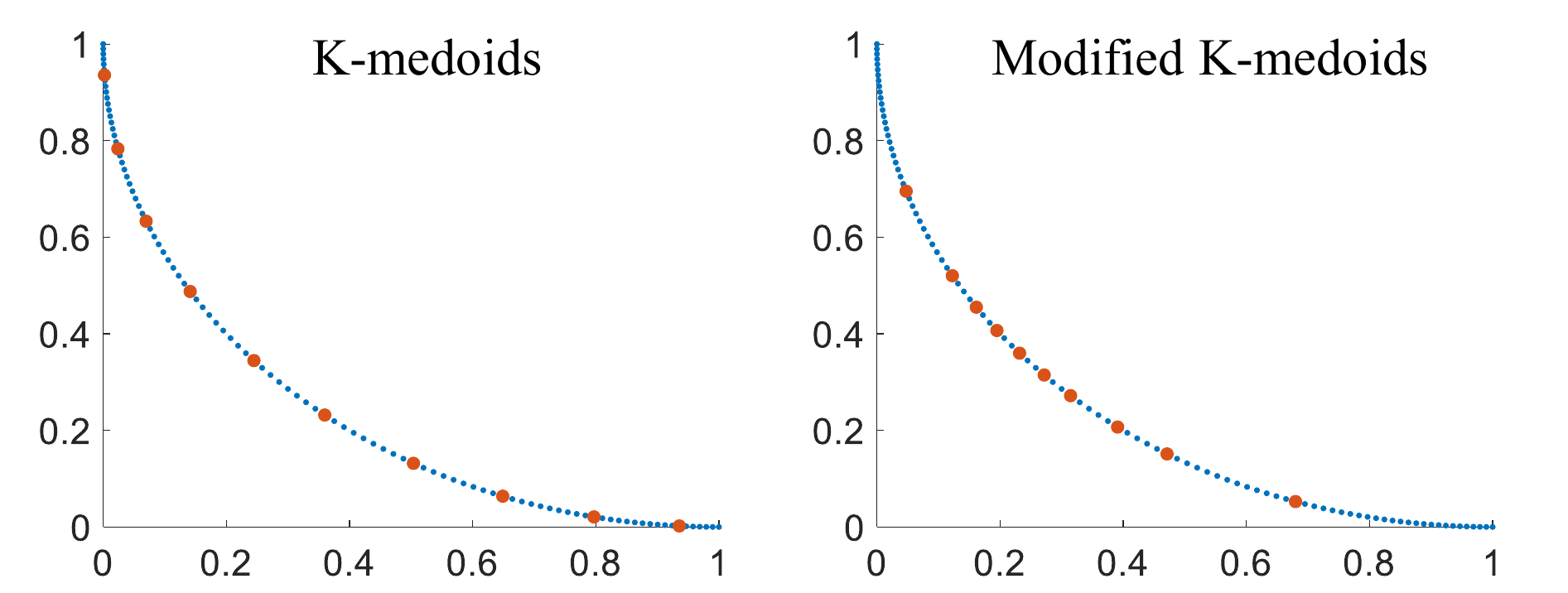}}
  \subfigure[Wave.] {\includegraphics[width=0.42\textwidth,trim=0 0 0 0,clip]{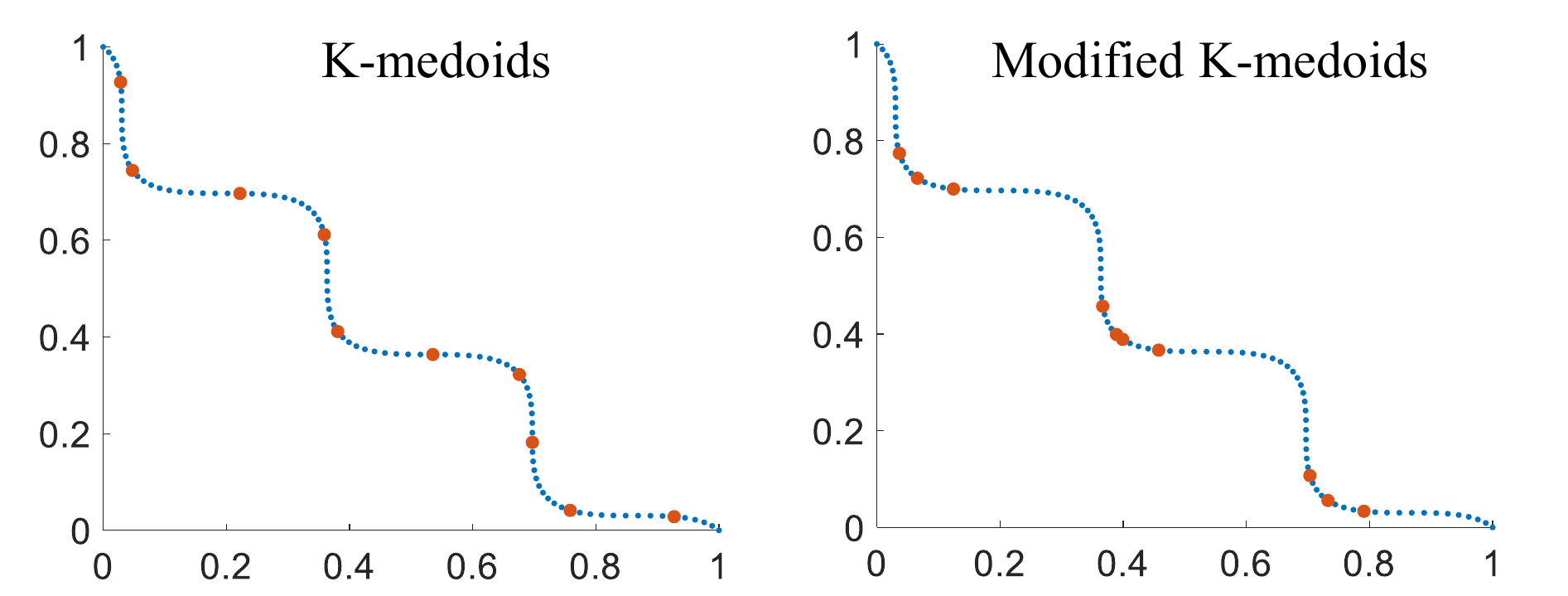}}
  \subfigure[DEB2DK.] {\includegraphics[width=0.42\textwidth,trim=0 0 0 0,clip]{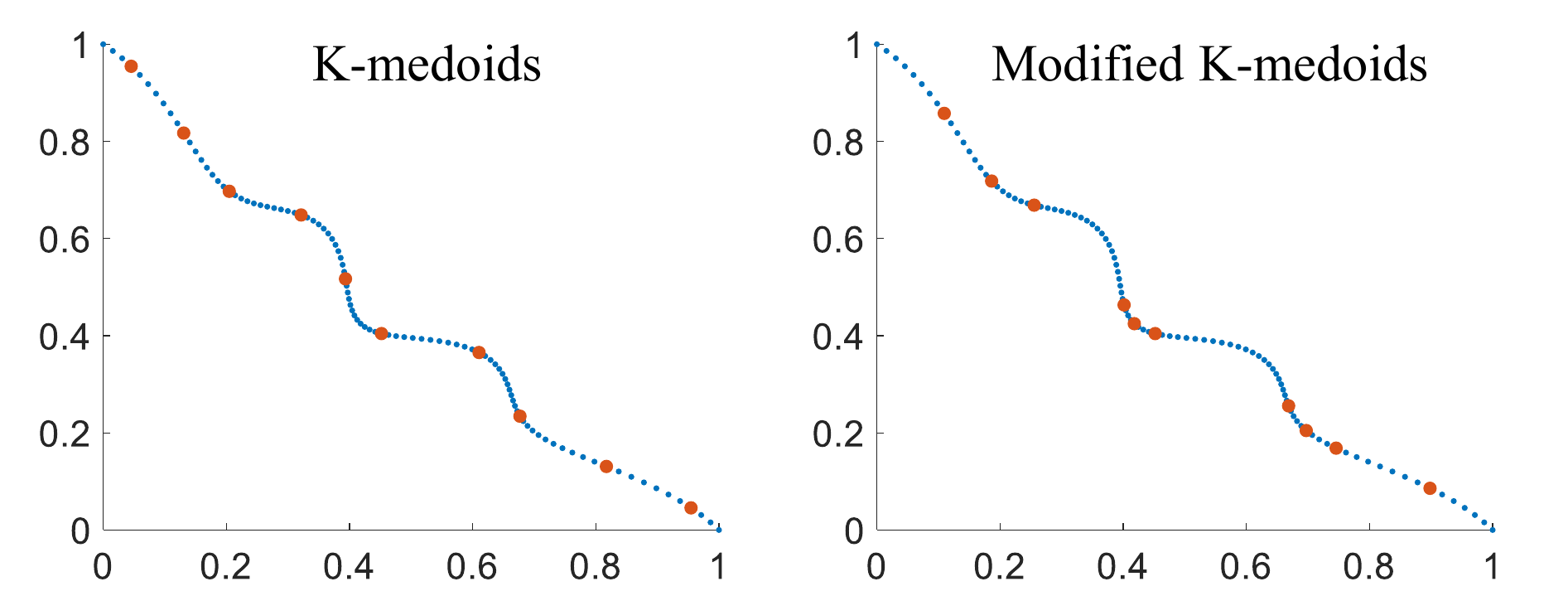}}
  
  \caption{Subsets selected by K-medoids and modified K-medoids on two-objective data sets.}
\end{figure*}

 \begin{figure*}[htbp]
\centering
  \subfigure[Linear.] {\includegraphics[width=0.42\textwidth,trim=0 0 0 0,clip]{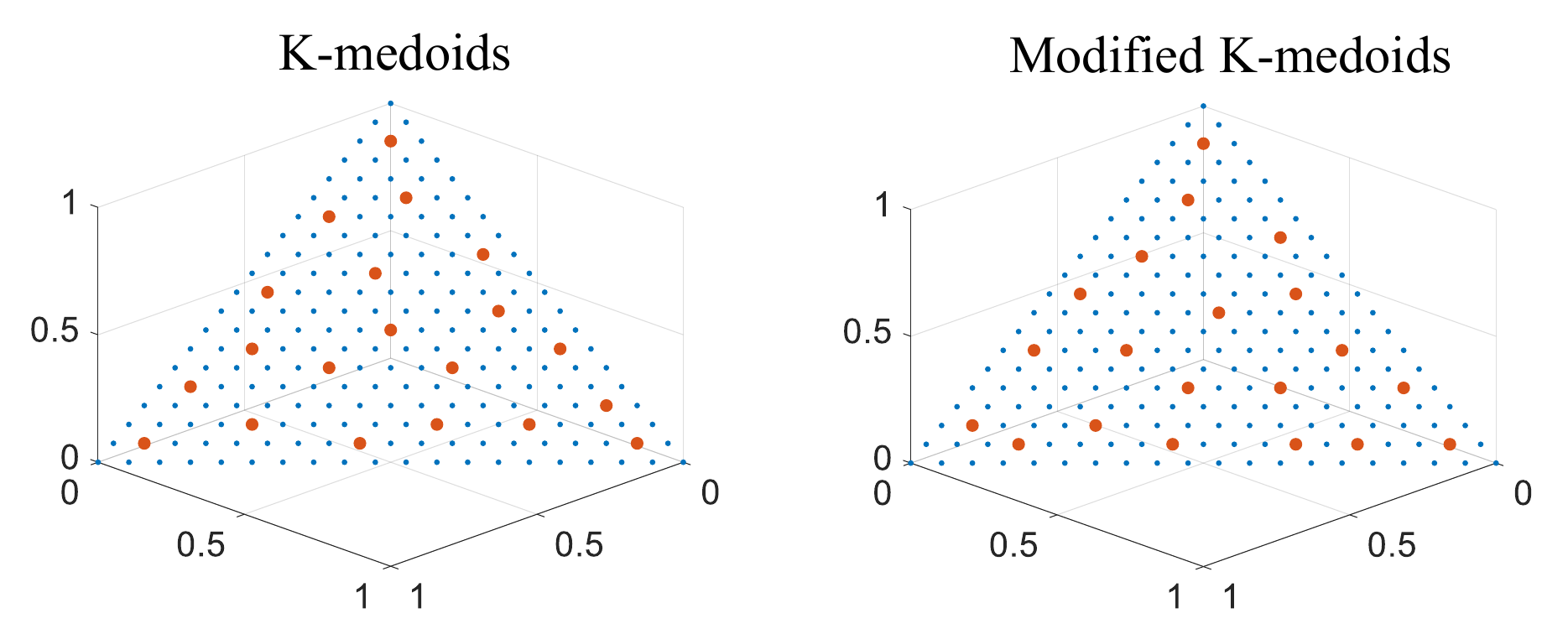}}
  \subfigure[Convex.] {\includegraphics[width=0.42\textwidth,trim=0 0 0 0,clip]{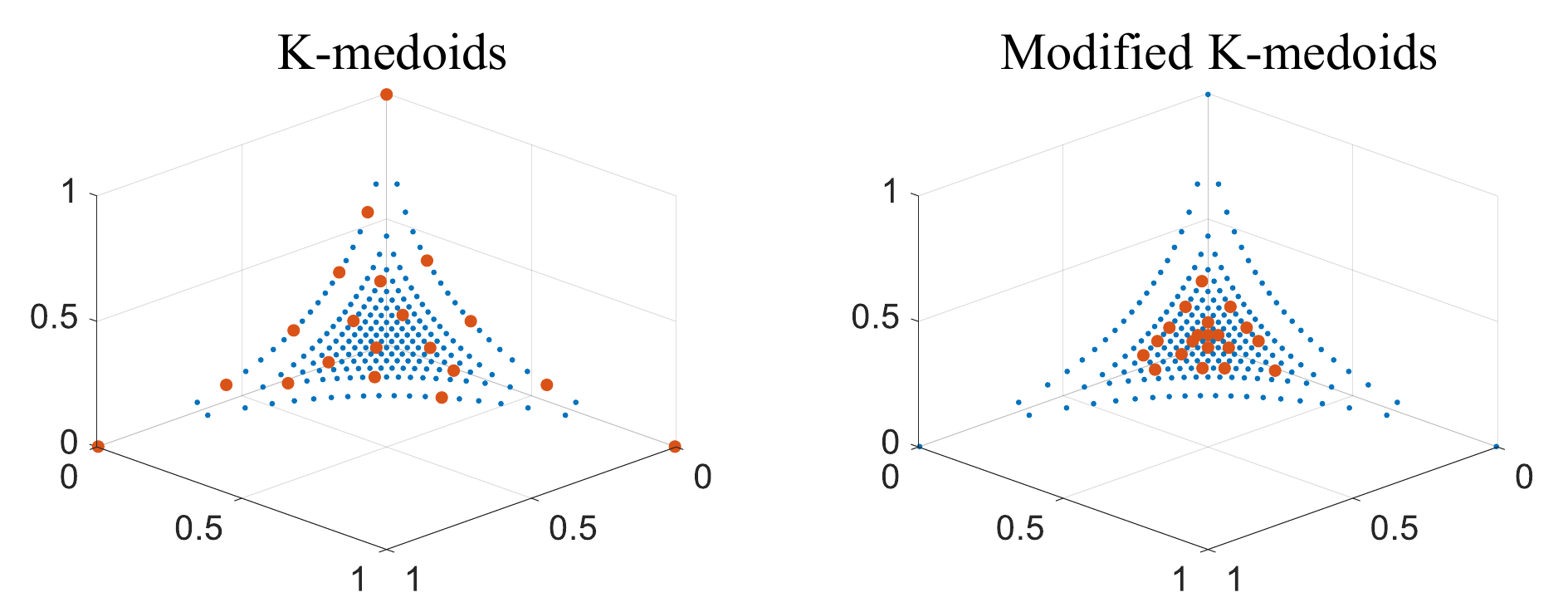}}
  \subfigure[Inverted Concave.] {\includegraphics[width=0.42\textwidth,trim=0 0 0 0,clip]{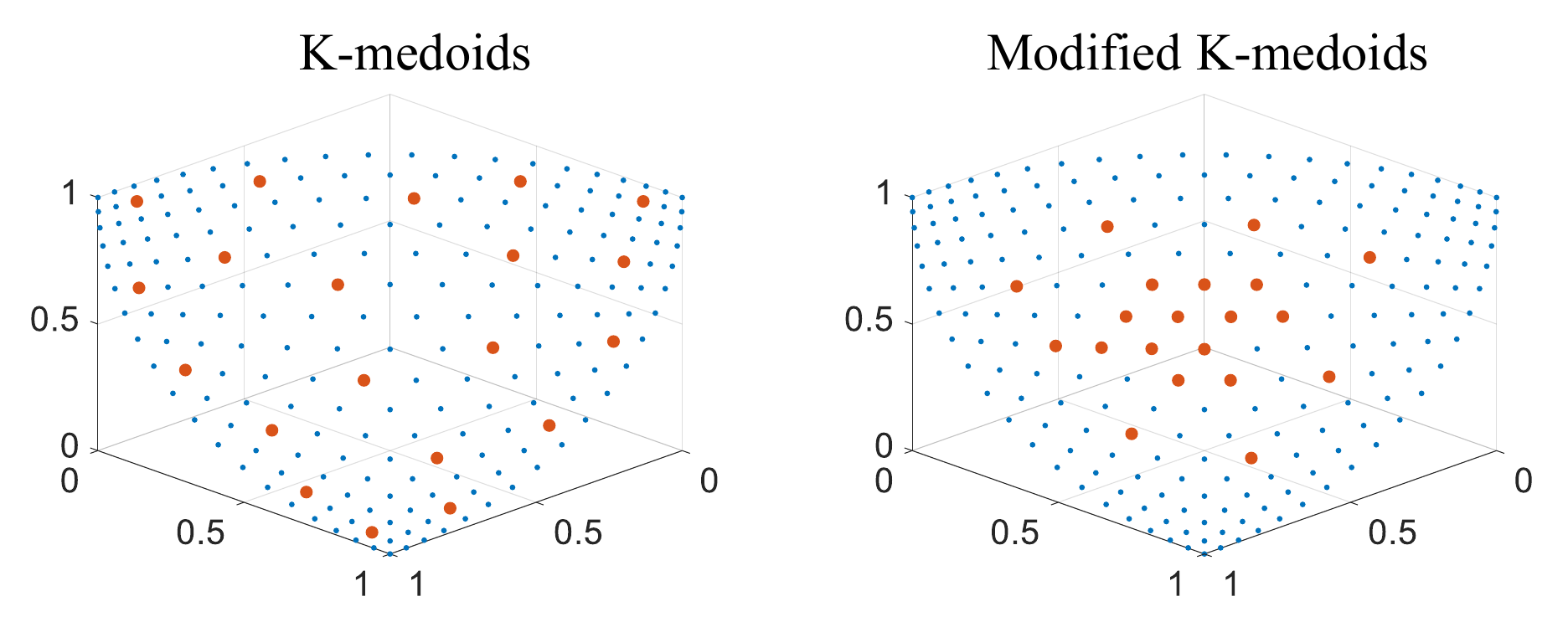}}
  \subfigure[DEB2DK.] {\includegraphics[width=0.42\textwidth,trim=0 0 0 0,clip]{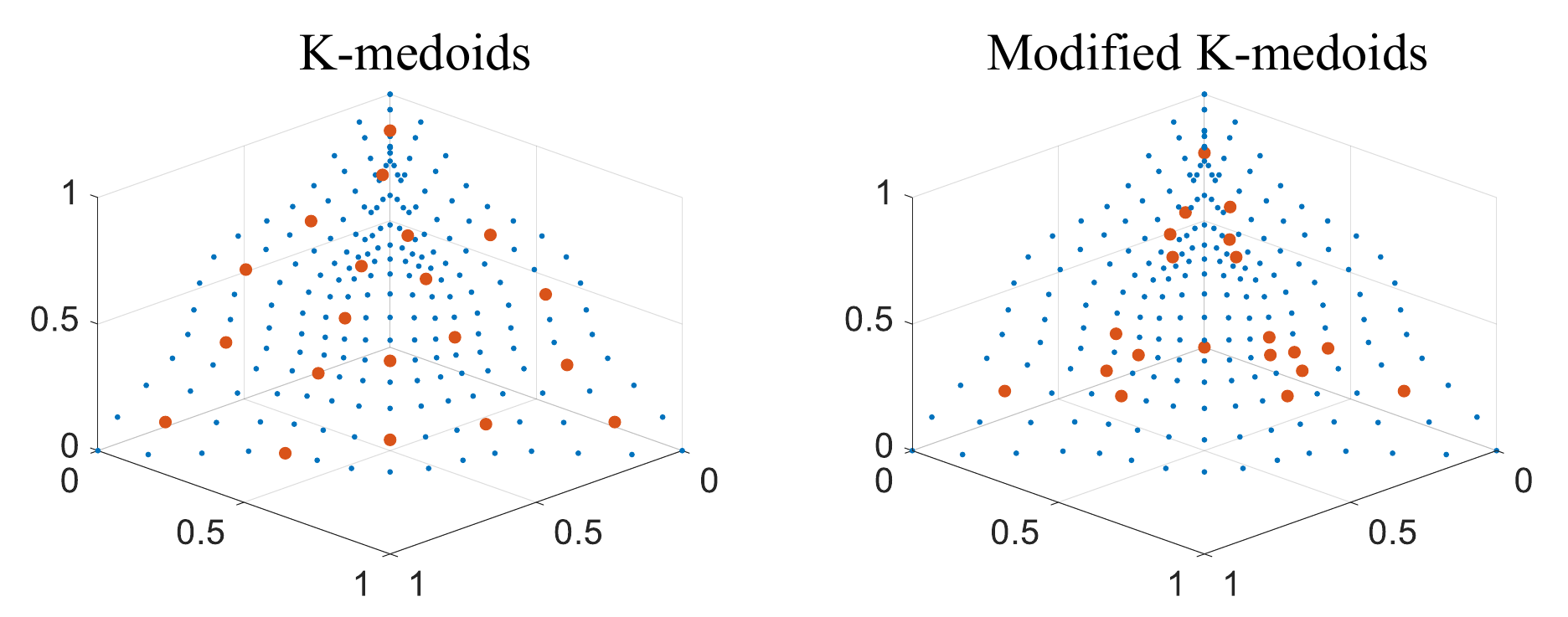}}
  
  \caption{Subsets selected by K-medoids and modified K-medoids on three-objective data sets.}
\end{figure*}

 Fig. 4 shows the subsets obtained by K-medoids and the modified K-medoids on two-objective data sets with 100 data points. The blue points in Fig. 4 are data sets and the red points are selected subsets. The number of clusters (i.e., subset size) is set to 10. We generate four data sets with various shapes. The linear and convex data sets contain uniformly distributed data points on the true Pareto front of DTLZ1 \cite{DTLZ} and MaF3 \cite{MaF}, respectively. The other two data sets (i.e., wave and DEB2K) with three knee points are generated according to \cite{Wave} and \cite{Knee2K}, respectively.

 We can see that for the linear data set, similar results are obtained from K-medoids and the modified K-medoids in Fig. 4 (a). However, for the convex data set in Fig. 4 (b), most of the points selected by the modified K-medoids are distributed in the knee region. In Fig. 4(c) and Fig. 4(d), we can see that knee regions are clearly identified by the modified K-medoids, and most of the selected solutions are in these regions.
 
 Fig. 5 shows subsets selected by K-medoids and modified K-medoids on three-objective data sets with 210 data points. Accordingly, the size of the subset is increased to 20. We can observe a similar result from Fig. 5 as in Fig. 4 (i.e., most solutions are in knee regions).
 These experiments clearly demonstrate that we can modify K-medoids to introduce decision-maker's preference for knee regions, which is important to decision-makers. Other clustering-based subset selection methods can also be modified in a similar way. 

\section{Conclusion and Future Work}
In this paper, we reviewed popular clustering algorithms and pointed out that IGD-based subset selection can be viewed as clustering. Then, we compared the IGD value of the obtained subset and the computation time of each algorithm on three-objective and eight-objective data sets. Our experimental results showed that when the data set size is small, hierarchical clustering with weighted average linkage is the best choice in the low-dimensional case and greedy IGD-based subset selection is the best choice in the high-dimensional case. When the data set size is large (e.g., subset selection from the UEA), K-means++ is the best in the low-dimensional case and greedy IGD-based subset selection is the best choice in the high-dimensional case. In addition, we showed that by modifying the similarity measure, we can include decision-maker's preference in clustering. 

As shown in this paper, general clustering algorithms have weakness in some scenarios since they are not specially designed for subset selection in EMO algorithms. Proposing clustering algorithms suitable for EMO algorithms is a promising future research direction. Additionally, comparing different clustering algorithms in more scenarios using various performance indicators is also an important future research topic. Clustering algorithms will be further utilized in EMO algorithms.





\bibliographystyle{IEEEtran}
\bibliography{ref}


\end{document}